\documentclass[nohyperref]{article}

\usepackage{microtype}
\usepackage{graphicx}
\usepackage{subfigure}
\usepackage{booktabs} 

\usepackage{hyperref}

\usepackage[accepted]{icml2022}

\usepackage{amsmath}
\usepackage{amssymb}
\usepackage{mathtools}
\usepackage{amsthm}

\usepackage[capitalize,noabbrev]{cleveref}

\theoremstyle{plain}

\theoremstyle{definition}

\theoremstyle{remark}

\usepackage[textsize=tiny]{todonotes}

\icmltitlerunning{Type theory in human-like learning and inference}

\begin{document}

\twocolumn[
\icmltitle{Type theory in human-like learning and inference}



\icmlsetsymbol{equal}{*}

\begin{icmlauthorlist}
\icmlauthor{Felix A. Sosa}{yyy}
\icmlauthor{Tomer D. Ullman}{yyy}
\end{icmlauthorlist}

\icmlaffiliation{yyy}{Department of Psychology, Harvard University, Cambridge, MA, United States}

\icmlcorrespondingauthor{Felix A. Sosa}{fsosa@fas.harvard.edu}

\icmlkeywords{Learning, Inference, Type Theory, Program Synthesis, Cognitive Modeling, ICML}

\vskip 0.3in
]



\printAffiliationsAndNotice{}  

\begin{abstract}

Humans can generate reasonable answers to novel queries \citep{schulz2012finding}: if I asked you what kind of food you want to eat for lunch, you would respond with a food, not a time. The thought that one would respond `after 4pm' to `what would you like to eat' is either a joke or a mistake, and seriously entertaining it as a lunch option would likely never happen in the first place. While understanding how people come up with new ideas, thoughts, explanations, and hypotheses that obey the basic constraints of a novel search space is of central importance to cognitive science, there is no agreed-on formal model for this kind of reasoning. We propose that a core component of any such reasoning system is a type theory: a formal imposition of structure on the kinds of computations an agent can perform, and how they're performed. We motivate this proposal with three empirical observations: adaptive constraints on learning and inference (i.e. generating reasonable hypotheses), how people draw distinctions between improbability and impossibility, and people's ability to reason about things at varying levels of abstraction.
\end{abstract}

\section{Introduction and Motivation}
Learning and inference are major focus points in cognitive science and artificial intelligence. Both learning and inference have been framed as inductive reasoning procedures, whereby agents figure out the latent causes that account for their observations. Recent formulations of learning and inference more specifically cast them as a program synthesis problem: inferring a generative program that best accounts for given observations \cite{gulwani2017program, bramley2018learning, rule2020child}. The program synthesis approach captures many core aspects of human-like learning and inference, such as generality, compositionality, and sample complexity \citep{lake2015human, ullman2020bayesian, lake2017building, piantadosi2012bootstrapping}. 

For all its attractive properties as a cognitive model, program synthesis is also hard. This is for two reasons: the space of programs is infinite, and the learning landscape over the space of programs is not smooth \cite{ullman2012theory}. If we are to take program synthesis seriously as a model of the mind, it is crucial to make program synthesis tractable in real-world settings.

Here, we review behavioral results that suggest people represent and use \textit{types} to constrain the problem of learning program-like models of the world. We propose that types are the basis for several intelligent behaviors, such as generating reasonable answers to novel queries, differentiating impossibilities from improbabilities, and reasoning about things at varying levels of abstraction. 

Types are usually viewed as annotations on programs that describe and enforce a program's expected behavior \cite{pierce2002types}. 

For example, consider the following Scheme program:
\begin{equation}
    \texttt{(define add x y) (+ x y)}
\end{equation} 
This program defines the function \texttt{add}, which takes in two arguments and adds them together. While to a person reading the program it may seem the program is only expected to take in numbers, nothing about the program itself makes that constraint known or enforces it. People can erroneously give non-numerical inputs to the program, and get nonsensical or unpredictable results. Anyone who has programmed in languages like javascript has likely experienced this.

We can annotate our Scheme program with a type signature, which denotes the type of input and  output expected by the program: 
\begin{equation}
    \begin{gathered}
        \texttt{(: add (Number -> Number))} \\
        \texttt{(define add x y) (+ x y)}
    \end{gathered}
\end{equation}

While the annotation seems simple, it imposes a rigid structure on the program's behavior: the program will only accept inputs of type \texttt{Number}, and will only return outputs of type \texttt{Number}. 

Now, consider giving a program-learning agent input-output pairs such as $\{([1,2],3), ([3,4], 7), ...\}$, and having it infer the program above, including the type-signature. This inferred program has the following, human-like properties: First, it can reject nonsensical input before carrying out further computation. If someone were to ask you ``what is 3 + dog?'', you might reasonably respond with a quizzical look, akin to encountering a type error. Second, the agent can restrict its space of possible traces to those given by the types used. If given the query `what input will generate the output `7'?', the search process can query over numbers, rather than over things like `dog', @, or an image of the pope. Similarly, humans seem to search sensibly in response to queries \cite{ullman2016coalescing}. Third, type signatures allow an agent to make predictions about the program, without having to go through an entire computation trace. For example, if you were asked to give the output of the program, given the inputs $\{x=4242, y=523525\}$, you can first say with certainty the output will be a number, even before you know what the exact number is.

We next consider empirical findings that support the idea that human and non-human animal inductive reasoning makes use of types. 

\section{Types strongly constrain inference and learning}

Both humans and non-human animals have inductive biases that make them more prepared to solve problems unique to their evolutionary history \cite{seligman1970generality}. One interpretation of the form of these inductive biases is that they mark what \textit{not} to think of when solving a problem or making a judgment \cite{phillips2019we}. Exactly how people know what not to think is not clear, yet \citealt{phillips2019we} provide an outline of a planning system that can capture this ability, by making planning a two-stage process in which an agent first comes up with a small action set with likely-valuable actions, and then runs a more model-based planning algorithm to examine each option in context. Our proposal differs from this outline, in that the first stage involves inferring a type-signature that defines the space of reasonable programs, and the second stage infers a likely-valuable program that inhabits that type. As we discuss below, these types act as inductive biases that carve out appropriate hypothesis spaces in learning and inference, conveying attractive learning dynamics such as zero- to few-shot learning in non-human animals, and the ability to generate reasonable answers to novel queries.

\subsection{Learning associations in animals}
There are examples of type constraints operating in simple learning problems in non-human animals: \citealt{garcia1966relation} showed what is known as the Garcia effect, or conditioned taste aversion, where animals will preferentially conclude that gustatory stimuli (e.g. water or food) are the cause of gustatory responses (e.g. nausea or stomach-related illness), even when there are other causes equally present during training (a bright light, a loud noise). In \citealt{garcia1966relation} rats were given either sugar water or water paired with a loud noise and bright light. After conditioning the rats to these two stimuli, the water was additionally paired with either an electric shock (causing pain) or radiation (causing nausea). Following the delivery of either the shock or the radiation, the rats that were given radiation avoided drinking the sweet water, whereas the rats that were shocked did not. Additionally, the rats that were shocked avoided the water with a loud noise and bright light, whereas the rats that were radiated did not. Notably, the avoidance of the sweet water after experiencing nausea was reproducible regardless of the time delay between drinking and experiencing nausea. This suggests that there is a structural bias that strongly constrains the associations animals will make even in simple learning paradigms, and that these constraints can be aptly described by types (e.g. gustatory stimuli uniquely explain gustatory responses).

\subsection{Generating hypotheses in humans and animals}

There are examples of typed hypothesis generation in adults and children: \cite{chu2021children} demonstrate that both adults and children (ages 4 to 8 years old) entertain novel, unverified claims to novel questions only when those claims can potentially answer the question, and judge claims that cannot answer the question as equally bad. For example, if you are told ``Sally applied to and wants to get into her favorite university. One day Sally went to her mailbox and found a letter. She read the letter and jumped up an down with excitement.'' and asked the question ``Why is Sally jumping up and down?'', you are likely to respond with reference to an acceptance letter. You likely wouldn't consider claims such as ``Sally wants to get into university'' or ``Sally has a favorite university'' as valid responses, because they don't answer the question, even though both of these claims are factual according to the story prompt and valid English sentences. \citealt{chu2021children} demonstrate claims that cannot answer questions are judged equally poorly by both adults and children, suggesting a discontinuity in the perceived value between potential answers and non-answers in both adults and children. 

Further, \cite{ullman2016coalescing} explore how adult  generate potential answers to novel queries, by asking people to come up with new names for particular entities, such as a pub. \citealt{ullman2016coalescing} demonstrate the answers people report all follow a rigid structure. For example, suppose someone asked you to come up with a new name for a pub. You may begin listing things such as `The Slug and Leaf' or `The Purple Duke'. You are unlikely to consider something like `Edgar Allen Poe loves eating a banana sandwich', though that is an acceptable sentence in English, and a reasonable answer to a different query. While that string of words can be rejected as unlikely once considered, the idea is that people do not even consider it in the first place.

The proposed model in \cite{ullman2016coalescing} samples a set of exemplars from a given category, and infers a syntax tree that best accounts for the overall structure of the examples. The model then instantiates the tree with terminal values (words, in this case), leading to novel names that correspond to the broad category that was observed. For example, after sampling some pub names, the model may infer the overall proposed trees `The [Adjective] [Noun]' and `The [Noun] and [Noun]'. Further soft restrictions on an embedding space of words keeps proposals within the semantic space of the examples. 

The model is simple, but captures a crucial aspect of how people might generate novel hypotheses: people may infer a template or schema that defines the expected structure of a solution to the problem, and use that template to guide further inferences about a solution. While the specific model uses syntax trees, the idea can be expanded to more general types, by expanding the task into the domain of inferring causal mechanisms in the form of programs. Here, the task would be, given a series of example program execution traces, infer a detailed type  (such as a refinement type, c.f. \citealt{polikarpova2016program}) and then generate novel programs that inhabit the type. This integrates well with the findings of \citealt{chu2021children} that adults and children judge non-answers as equally bad; in this case adults and children are both recognizing the non-answer cannot inhabit the inferred type of the expected solution and discard it without further investigation.


These studies on humans and non-human animals show that humans readily generate reasonable answers to new questions and that animals associate certain cues more naturally with certain responses. The computational machinery used to account for these results in humans an non-human animals has often made use of various biases. Instead, we find it more plausible and useful to group certain cues and outcomes as belonging to the same type.

\section{Improbability and impossibility}
People distinguish impossible events from improbable events. For example, \cite{shtulman2007improbable} presented children and adults with events that were either ordinary (e.g. eating vegetables for dinner), improbable (e.g. finding an alligator underneath your bed), and impossible (e.g. eating lightning for dinner). Participants categorized the events into either of the three categories. Children and adults both found the same things impossible and the same things ordinary. Young children classified improbable events as impossible compared to adults, but this difference decreased with age.

These findings suggest that distinctions between possibility and probability are present relatively early, though they are refined through experience. We suggest that some kind of impossibility judgements come about from a type system akin to innate systems of Core Knowledge \citep{spelke2007core}. This distinction is further evidenced by neural signatures of surprise to sentence comprehension, which we take as neural evidence for a distinction between improbable and impossible in humans.

Some of the neural evidence for a distinction between low probability and truly impossible events comes from EEG studies. When viewing surprising or unexpected stimuli in a controlled setting, a positive electric potential, known as the P300, is often observed 300ms-600ms after viewing the surprising stimuli \cite{duncan1982p300, levi2020surprise}. Another neural signal similar to the P300, known as the N400, has been observed in sentence comprehension tasks, and is associated with a different kind of surprise that aligns well with improbability. Specifically, people will elicit an N400 signal if they see a surprising word in a sentence that violates the expected semantics of the sentence (e.g. ``He spread the warm bread with \textit{socks}'') \cite{kutas1980reading}. We take these results as neural evidence that people have distinct notions of surprise, and interpret them as distinctions between improbable (e.g. pulling a red marble out of a bag of 100 white marbles and 1 red marble), and impossible (e.g. pulling a hamburger out of the same bag). We suggest that more research is needed to compare these with inconceivability (e.g. being asked 'Can you add the concept of the number 2 to a hamburger?'). 

A possible rebuttal to our type theory proposal is that impossible and inconceivable events exist in the same continuum as all other events: impossible events are simply zero probability events, or extremely low probability events. But this formulation misses a key distinction between improbability and impossibility, and an important subtlety about the different ways in which something can be impossible: impossible events may be zero-probability events, or they may be invalid under the assumptions of the structure of how our world works (e.g. time traveling), or they may be inconceivable, nonsensical, or illogical (e.g. a circle that is a square). Such concepts have been debated in philosophy \cite{gendler2002conceivability}, and we take them to be discrete categories that are better explained by type-theory than by models that only traffic in probabilities, such as large-language models.
Under this interpretation, there are impossible events that simply ``don't compute''. Under the analogy of program synthesis, these would be programs that can't evaluate (i.e. they have no normal form); impossible events or entities of this kind would be events who have no well-typed program instance that can define the event or entity.

\section{Abstraction}
By representing types, one can reason abstractly about the behavior of a space of hypotheses in expectation, without having to consider individual hypotheses in the space. One final empirical finding that motivates the suggestion of typed computation in the mind is the ability people have to reason at varying degrees of abstraction. While there is plenty of work demonstrating how people use abstract representations to perform planning tasks \citep{sanborn2018representational,ho2021control}, our concerns are demonstrated with the following example: Suppose I ask you whether a tennis ball can go through a solid brick. You can answer this using different mental computations. Perhaps you create a mental simulation of a ball and a wall, and run that simulation until the two collide \cite{battaglia2013simulation}. Or perhaps you use deductive logical reasoning, something such as `the ball is solid, the wall is solid, solid things cannot go through one another, so no' \cite{forbus1988qualitative}. I can ask you this question at varying levels of commitment to detail: a ball, a tennis ball, a blue 10kg tennis ball that is 30cm in diameter. Each of these representations varies in terms of their commitment to detail, but each of them obeys a set of constraints by themselves being physical objects in the shape of a sphere. The ability to preferentially reason about something abstractly, concretely, or somewhere in between in an ability
that people use everyday to plan and solve various tasks like walking to work or figuring out how to make amends with a close friend.

We suggest that types readily capture this ability to adaptively commit to different representations of a solution to a problem if we allow the types themselves to be computable, as is the case in Computational Type Theory \cite{angiuli2016computational}). For example, consider again the adding program in Eq. 2.
We can take just the type signature \texttt{(: add (Number -> Number))}, and evaluate it on the input \texttt{Number}, in which case it returns \texttt{Number}. This value can be used to do further computations, in the same way that we would compose the output of any program with another program. This offers us a formal framework for evaluating programs at differing levels of abstraction. In a more human-meaningful example, we could imagine the task of figuring out how to get to our favorite park from home. Rather than having to simulate all possible trajectories in full detail (e.g. the texture of the sidewalk, color of the signs, etc.), we only need to pay attention to high-level structures that have meaningful values for constraining our hypothesis space over trajectories, such as the general direction we need to walk in, the distance we might walk, etc. Each of these high-level structures can be modeled as a placeholder for hypothesis spaces, but themselves can be evaluated (e.g. knowing I need to go North indeed does computational work for me by limiting the space of directions I can walk to just those that are North). This finds a natural description as a type, in the same way \texttt{Number} describes a body of mathematical objects with specific properties, so can the type of \texttt{Direction} or \texttt{Location}.

\section{Discussion}
We suggested that people should and do take advantage of type-level annotations to perform inductive reasoning. We discussed how types can give an agent strong inductive biases that constrain learning and inference. We considered how types can model the differentiation between impossibility from improbability, and enable agents to reason at varying levels of abstraction about a representation. We reviewed empirical findings that can be re-cast as evidence for our proposal. We note that this proposal is speculative, and needs both empirical investigation in people an non-human animals, as well as formal specification. That is, the proposal is specified at a high level of abstraction, but still needs a formal account of a type theory in a cognitively relevant domain, such as intuitive physics \cite{kubricht2017intuitive} or psychology \cite{baker2012bayesian}, and an empirical experiment that demonstrates qualitative and quantitative evidence for people paying attention to and using types to reason about a cognitively relevant domain. We have suggestions for each of these missing components, and are interested in discussing them with the workshop community. While we are well-aware of the limitations of the current proposal, we are confident that it can spark lively conversation about the nature of current models of reasoning, and foster collaborations between investigators in programming languages and cognitive science.


\bibliography{example_paper}
\bibliographystyle{icml2022}


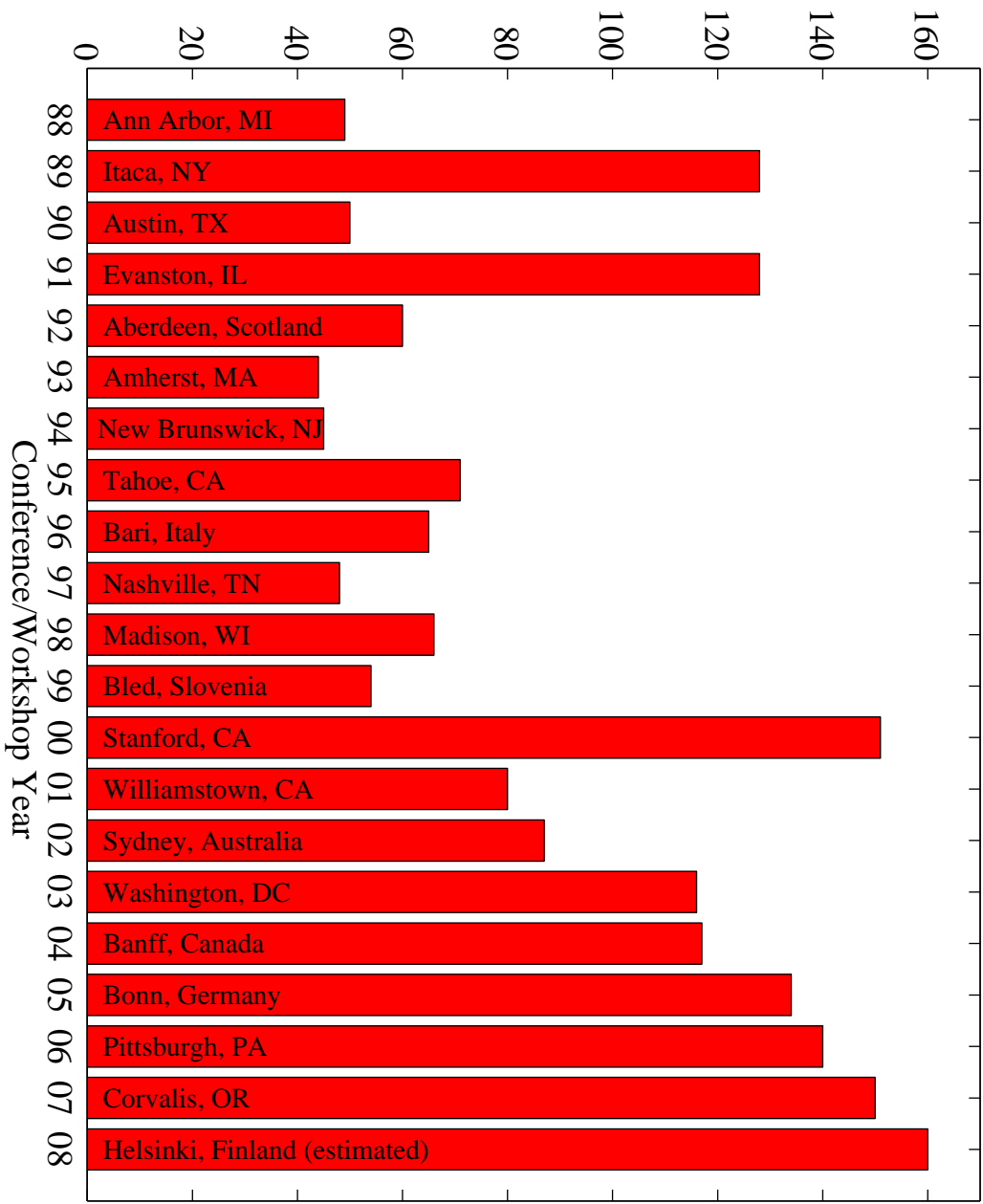
\end{document}